\documentclass[a4paper,fleqn]{cas-sc}

\usepackage[numbers]{natbib}
\usepackage{subfigure}  
\usepackage{algorithmicx}  
\usepackage{algpseudocode}  
\usepackage{amsmath}
\usepackage{algorithm}    
\newcommand{\tabincell}[2]{\begin{tabular}{@{}#1@{}}#2\end{tabular}}

\makeatletter
\DeclareRobustCommand\onedot{\futurelet\@let@token\@onedot}
\def\@onedot{\ifx\@let@token.\else.\null\fi\xspace}
\def\eg{\emph{e.g}\onedot} 
\def\ie{\emph{i.e}\onedot} 
 
\def\etc{\emph{etc}\onedot} 
 
\def\etal{\emph{et al}\onedot}

\usepackage{xspace}
\newcommand{\name}[0]{PAMT\xspace}

\usepackage{xcolor}

\def\tsc#1{\csdef{#1}{\textsc{\lowercase{#1}}\xspace}}
\tsc{WGM}
\tsc{QE}
\tsc{EP}
\tsc{PMS}
\tsc{BEC}
\tsc{DE}

\begin{document}
\let\WriteBookmarks\relax
\def\floatpagepagefraction{1}
\def\textpagefraction{.001}

\shorttitle{PAMT for Node Classification on Attributed Networks}

\shortauthors{J. Chen et~al.}

\title [mode = title]{Propagation with Adaptive Mask then Training for Node Classification on Attributed Networks}      

%
\author[1,2]{Jinsong Chen}[style=chinese, orcid=0000-0001-7588-6713]


\fnmark[1]

\ead{chenjinsong@hust.edu.cn}



\affiliation[1]{organization={School of Computer Science and Technology, Huazhong University of Science and Technology},
    city={Wuhan, Hubei},
    postcode={430074}, 
    country={China}}

\affiliation[2]{organization={Institute of Artificial Intelligence, Huazhong University of Science and Technology},
    city={Wuhan, Hubei},
    postcode={430074}, 
    country={China}}

\author[1]{Boyu Li}[style=chinese]
\ead{afterslby@hust.edu.cn}
\fnmark[1]
\author[1]{Qiuting He}[style=chinese]
\ead{heqiuting@hust.edu.cn}

\author[1]{Kun He}[style=chinese]
\cormark[1]
\ead{brooklet60@hust.edu.cn}

\cortext[cor1]{Corresponding author.}
\fntext[fn1]{Co-first authors.}

\begin{abstract}
Node classification on attributed networks is a semi-supervised task that is crucial for network analysis. 
By decoupling two critical operations in Graph Convolutional Networks (GCNs), namely feature transformation and neighborhood aggregation, some recent works of decoupled GCNs could support the information to propagate deeper and achieve advanced performance.
However, they follow the traditional structure-aware propagation strategy of GCNs, making 
it hard to capture the attribute correlation of nodes and sensitive to the structure noise described by edges whose two endpoints belong to different categories. To address these issues, we propose a new method called the {\itshape Propagation with Adaptive Mask then Training} (\name). 
The key idea is to integrate the attribute similarity mask into the structure-aware propagation process. In this way, \name could preserve the attribute correlation of adjacent nodes during the propagation and effectively reduce the influence of structure noise.
Moreover, we develop an iterative refinement mechanism to update the similarity mask during the training process for improving the training performance.
Extensive experiments on four real-world datasets demonstrate the superior performance and robustness of \name.
\end{abstract}

\begin{highlights}
\item We propose a new semi-supervised method for the node classification on attributed networks. 
\item We propagate the label information on the weighted graph by combining attribute feature and topology feature.
\item We develop an iterative refinement to boost training and use a momentum-based strategy for a stable training.
\item Extensive experiments demonstrate the superiority on performance and the robustness to structure noise.
\end{highlights}


\maketitle

\section{Introduction}
Attributed networks have been widely used to model various entities and their relationships in the real world, such as users and their social connections in social networks. 
Due to the complex topology structure and abundant attribute features, graph mining tasks on attributed networks are much more challenging. 
In this work, we address a fundamental and essential task, the semi-supervised node classification (SSNC) on attributed networks, which has a wide range of applications such as social influence analysis in social networks~\cite{qiu2018deepinf,sonet}.

Numerous techniques have been proposed for SSNC, \eg, network embedding-based methods~\cite{SANE} and label propagation-based methods~\cite{xie2021graphhop}, \etc. Among which, Graph Convolutional Network (GCN) based methods~\cite{DBLP:conf/nips/HamiltonYL17,DBLP:conf/iclr/KipfW17,DBLP:conf/iclr/VelickovicCCRLB18,DBLP:conf/icml/XuLTSKJ18} have gained great success with high performance.
The core component of GCN-based models is the graph convolution layer.
Each graph convolution layer includes two important operations, namely neighbor aggregation and feature transformation. 
Neighbor aggregation is used for aggregating features from the neighborhood to update node representations, while the function of feature transformation is to extract new representations from the previous features~\cite{DBLP:conf/www/DongCFHBD021}. 
These operations are actually coupled in a graph convolution layer for aggregating and transferring the information of neighborhood nodes to learn the node representations.

Despite effectiveness, recent studies indicate that the coupled design of GCN could lead to some problems in practice, such as over-smoothing~\cite{DBLP:conf/aaai/ChenLLLZS20,DBLP:conf/kdd/LiuGJ20} and inefficient training~\cite{DBLP:conf/iclr/KlicperaBG19,DBLP:conf/icml/WuSZFYW19}.
A natural solution to the above issues is to decouple the operations of neighbor aggregation and feature transformation. Following this strategy, several effective models~\cite{DBLP:conf/iclr/KlicperaBG19,DBLP:conf/icml/WuSZFYW19} have been proposed, termed decoupled GCNs~\cite{DBLP:conf/www/DongCFHBD021}. 
The rationale of decoupled GCNs is to remove the nonlinear activation functions and collapse the matrices (adjacency matrix and parameter matrices) from the graph convolution layers. In this way, the neighbor aggregation and feature transformation can be considered as two separate modules. 
Such decoupled design could permit deeper propagation without leading to over-smoothing and exhibits superior performance for SSNC on attributed networks~\cite{DBLP:conf/iclr/KlicperaBG19}.    

Although decoupled GCNs simplify the design of GCN-based models and achieve significant improvements in terms of accuracy and efficiency, we observe that they still follow the traditional structure-aware propagation mechanism of GCN-based methods, \ie, only utilizing the adjacency matrix to aggregate information. 
Thus, they suffer from the following limitations on attributed networks:
\begin{enumerate}
\item \textbf{It is hard to capture the correlation of attribute features.} 
The attributed information is one of the essential features in attributed networks. Nodes with similar attributes tend to belong to the same category. 
However, existing propagation mechanisms of decoupled GCNs are only topology structure-aware and ignore the attribute features of nodes. 
Hence, the model is hard to capture the correlation of attribute features between nodes on attributed networks.   
In addition, recent studies have shown that the topology structure-aware propagation mechanism could destroy the attribute information of nodes~\cite{DBLP:conf/wsdm/JinDW0LT21,wang2020gcn}.
Thus, the topology structure-aware propagation mechanism would limit the model performance for node classification on attributed networks.

\item \textbf{They are sensitive to the graph structure noise.} 
The graph structure noise is described by edges connecting nodes of different categories~\cite{DBLP:conf/www/DongCFHBD021}. 
During the propagation process, this topology structure-aware propagation mechanism assigns large weights to the adjacent nodes no matter whether they belong to the same category, which will hurt the performance, causing the model to be sensitive to the graph structure noise, as demonstrated in recent works~\cite{DBLP:conf/www/DongCFHBD021}.
\end{enumerate}

To address the above issues, in this work, we propose a novel method called {\itshape Propagation with Adaptive Mask then Training} (\name). \name is inspired by~\cite{DBLP:conf/www/DongCFHBD021} which improves the training efficiency of decoupled GCNs by introducing a two-step training framework. 
The key idea is to integrate the attribute similarity of adjacent nodes into the structure-aware propagation process, so that the attribute correlation of the adjacent nodes could be well captured and the influence of structure noise could be effectively reduced.  

There are three main stages in the proposed \name:
1) It generates a similarity mask based on attribute features and labeled data to represent the similarity of each node pair.
2) A propagation matrix is generated based on the similarity mask and adjacency matrix, and the matrix is further used in the label propagation procedure to create soft labels for training the classifier.     
3) \name enhances the training process by using an iterative refinement mechanism to alternately update the similarity mask and the soft labels. In addition, a momentum-based update strategy~\cite{he2020momentum} is applied in the process of generating new soft labels so as to gain a stable training. 

Our main contributions are summarized as follows:

\begin{itemize}
	\item We propose a new method called \name for node classification on attributed networks, whose key idea is to utilize the attribute information to strengthen the existing propagation operation of decoupled GCNs. 
	\item We develop an iterative refinement mechanism applied in the training process to improve the training performance, and utilize a momentum-based update strategy to gain a stable training.
	\item Extensive experiments on four real-world datasets demonstrate the superiority on performance and robustness to structure noise of \name over the baselines.         
\end{itemize}

\section{Related Work}
In this section, we review related works on GCNs and decoupled GCNs, and highlight the difference of our method.

\subsection{GCN-based Approaches}
Many follow-up methods have been proposed recently to improve the efficiency of the vanilla GCN~\cite{DBLP:conf/iclr/KipfW17}.  GraphSAGE~\cite{DBLP:conf/nips/HamiltonYL17} develops a sampling-based aggregation operation to overcome the memory overflow issue of GCN when the graph is in large scale.  GAT~\cite{DBLP:conf/iclr/VelickovicCCRLB18} introduces an attention mechanism into the aggregation operation to capture the feature similarity between adjacent nodes, that are overlooked in the vanilla GCN. 
For improving the flexibility of aggregation operation of GCN, Xu \etal~\cite{DBLP:conf/icml/XuLTSKJ18} propose JK-Nets to selectively exploit information from neighborhoods of different locality.  
GCN suffers from the over-smoothing problem~\cite{DBLP:conf/aaai/ChenLLLZS20} when the graph convolutional layers get deeper, and a possible solution is to balance the information obtained between the layers~\cite{DBLP:conf/icml/Li0GK21}. 
Li \etal~\cite{DBLP:journals/corr/abs-2006-07739} propose DeeperGCN by using a generalized aggregation function to improve the training performance. 

\textbf{Integrating attribute information.}
Besides GAT~\cite{DBLP:conf/iclr/VelickovicCCRLB18} that utilizes attribute information to guide the neighborhood aggregation,
there are several GCN-based studies that combine attribute information and topology information for achieving high performance.
One common way is to generate a $k$-nearest neighbor ($k$NN) graph based on the attribute similarity matrix of nodes.
To obtain a similarity matrix, various methods have been adopted to calculate the similarity of nodes according to their attribute features, such as Cosine Similarity~\cite{AMGCN,DBLP:conf/wsdm/JinDW0LT21,DAG} and Heat Kernel~\cite{AMGCN}.
In addition to the $k$NN graph, recent studies utilize the attribute information to estimate the label similarity matrix~\cite{HOG} or the block matrix~\cite{BMGCN} and further combine them with the adjacency matrix for enhancing the effort of aggregation operation. 
For more details on GCN-based approaches, we refer the readers to a survey~\cite{wu2020comprehensive}. 

Different from the above GCN-based methods, the proposed \name adopts the decoupled design that uses independent modules for information propagation and feature transformation. \name also develops an iterative refinement mechanism to dynamically adjust the attribute similarity mask during the training process which makes a great difference from the above methods.  

\subsection{Decoupled GCN-based Approaches}
The two important operations of GCN, neighbor aggregation and feature transformation, are coupled in the graph convolution layer in the original design. Recent studies~\cite{DBLP:conf/iclr/KlicperaBG19,DBLP:conf/kdd/LiuGJ20,DBLP:conf/sigir/0001DWLZ020,DBLP:conf/icml/WuSZFYW19} have demonstrated that such coupled design may limit the model's capability of deeply capturing the topology feature and cause over-smoothing issue. 

To tackle the above issues, Wu \etal~\cite{DBLP:conf/icml/WuSZFYW19} propose SGCN by removing the nonlinear activation functions and collapsing weight matrices between consecutive graph convolutional layers. APPNP~\cite{DBLP:conf/iclr/KlicperaBG19} combines the personalized PageRank and the neural network classifier to simplify GCN by treating the neighbor aggregation and feature transformation as independent modules.
Dong \etal~\cite{DBLP:conf/www/DongCFHBD021} summarize studies based on the idea of decoupling neighbor aggregation and feature transformation, and call these methods the decoupled GCNs. The authors theoretically analyze the relationship between decoupled GCNs and label propagation (LP)~\cite{xiaojin2002learning}, and propose a Propagation then Training (PT) method and an improved version, Propagation then Training Adaptively (PTA). PT generates the soft labels for unlabeled nodes and trains the neural network classifier using the obtained labels in a supervised learning manner. PTA extends PT by introducing an adaptive weighting strategy into the training stage to further improve the model performance.

We observe that these methods ignore the attribute information during the propagation, causing them sensitive to the structure noise. To this end, our \name utilizes not only topology features but also attribute features when propagating information on networks, and improves the model robustness.

\section{Preliminaries}
In this section, we first formulate the task of semi-supervised node classification on attributed networks, then we briefly review the vanilla GCN and decoupled GCNs. Finally, we introduce the PT method~\cite{DBLP:conf/www/DongCFHBD021} in detail, which is highly related to our work.

\subsection{Problem Formulation}
Consider an unweighted and undirected attributed graph $G=(V, E, \mathbf{X})$, where $V$ and $E$ are the sets of nodes and edges, respectively. $n=|V|$ denotes the number of nodes. $\mathbf{X}\in \mathbb{R}^{n \times d}$ is the feature matrix of node attributes, where $d$ is the dimension of each feature vector. Let $\mathbf{A}$ be the adjacency matrix, \ie, $\mathbf{A}_{ij}=1$ if there exists an edge between nodes $v_i$ and $v_j$, otherwise, $\mathbf{A}_{ij} = 0$. Given a small set of labeled nodes denoted by $V_L$, the task of semi-supervised node classification is to predict the labels for all $v \in V \backslash V_L$ .

\subsection{Vanilla GCN and Decoupled GCNs}
The rationale of the vanilla GCN is to simultaneously capture the topological information of the graph and the attribute features of nodes by stacking several graph convolution layers. The graph convolution layer can be written as~\cite{DBLP:conf/iclr/KipfW17}:
\begin{equation}
	\mathbf{H}^{(l+1)} =\boldsymbol{\sigma}(\hat{\mathbf{A}}\mathbf{H}^{(l)}\mathbf{W}^{(l)}),
	\label{eq:GCN}
\end{equation} 
where $\hat{\mathbf{A}} = \widetilde{\mathbf{D}}^{-\frac{1}{2}}\widetilde{\mathbf{A}}\widetilde{\mathbf{D}}^{-\frac{1}{2}}$ is the normalized adjacency matrix, $\widetilde{\mathbf{A}}$ is the adjacency matrix with a self loop on each node, and $\widetilde{\mathbf{D}}$ denotes the corresponding diagonal degree matrix, \ie, $\widetilde{\mathbf{D}}_{ii} = \sum_{j} \widetilde{\mathbf{A}}_{ij}$.  $\mathbf{H}^{(l)}$ is the feature matrix of the $l^{th}$ GCN layer, and $\mathbf{H}^{(0)} = \mathbf{X} \cdot \mathbf{W}^{(l)}$ represents the learnable weight matrix of the $l^{th}$ GCN layer, and $\boldsymbol{\sigma}$ denotes the activation function. 

According to \autoref{eq:GCN}, there are two coupled operations in each graph convolution layer, namely neighbor aggregation and feature transformation. However, such coupled design limits the capability of deeply capturing the topology information and easily leads to over-smoothing problem. A rational solution is to decouple the two operations, termed the decoupled GCNs~\cite{DBLP:conf/www/DongCFHBD021}:
\begin{equation}
	\hat{\mathbf{Y}} = softmax(\bar{\mathbf{A}}\boldsymbol{f}_{\theta}(\mathbf{X})),
	\label{eq:dGCN}
\end{equation}       
where $\boldsymbol{f}_{\theta}$ indicates the neural network classifier (\eg, MLP~\cite{hornik1991approximation}), and $\bar{\mathbf{A}}=\sum \varphi_k \hat{\mathbf{A}}^k$ denotes the operation of propagating $k$ times through $\hat{\mathbf{A}}$ by a propagation strategy (\eg, personalized PageRank in APPNP). Each entry $\bar{\mathbf{A}}_{ij}$ denotes the proximity of nodes $i$ and $j$ in the graph. It has been demonstrated that such decoupled design can improve the performance of node classification~\cite{DBLP:conf/www/DongCFHBD021,DBLP:conf/iclr/KlicperaBG19}.       

\subsection{The PT Method}
As shown in \autoref{eq:dGCN}, the decoupled GCNs can be regarded as a combination of propagation module and classifier module. 
Based on such design, Dong \etal~\cite{DBLP:conf/www/DongCFHBD021} propose a novel method called Propagation then Training (PT). PT contains two main stages, \ie, the training stage and the inference stage. In the training stage, PT generates soft labels by label propagation, and trains the neural network classifier $\boldsymbol{f}_{\theta}$ based on the soft labels. The objective function is as follows:
\begin{equation}
\begin{aligned}
	\boldsymbol{L}_{PT}(\theta)  & = \boldsymbol{\ell}(\boldsymbol{f}_{\theta}(\mathbf{X}), \mathbf{Y}_{soft}),
	\label{eq:PT_LOSS}
\end{aligned}
\end{equation}        
where $\mathbf{Y}_{soft} = \bar{\mathbf{A}}\mathbf{Y}_{L}$ indicates the soft labels generated by $\bar{\mathbf{A}}$ and the observed labels $\mathbf{Y}_{L}$, and $\boldsymbol{\ell}(\cdot)$ denotes the cross entropy loss. 
In the inference stage, PT generates the final prediction for nodes as follows:
\begin{equation}
	\hat{\mathbf{Y}} = \bar{\mathbf{A}}\cdot softmax(\boldsymbol{f}_{\theta}(\mathbf{X})).
	\label{eq:PT_OUT}
\end{equation}
It indicates that for each node, PT first generates the soft labels according to the final output of the neural network classifier after softmax. Then, PT aggregates the soft labels of its $k$-hop neighbors (if propagate $k$ times) based on $\bar{\mathbf{A}}$. Finally, PT outputs the final prediction labels for each node by aggregating information of the neighbors.

\section{The Proposed \name}\label{sec3}
\begin{figure}[t]
\centering
	\includegraphics[width=15cm]{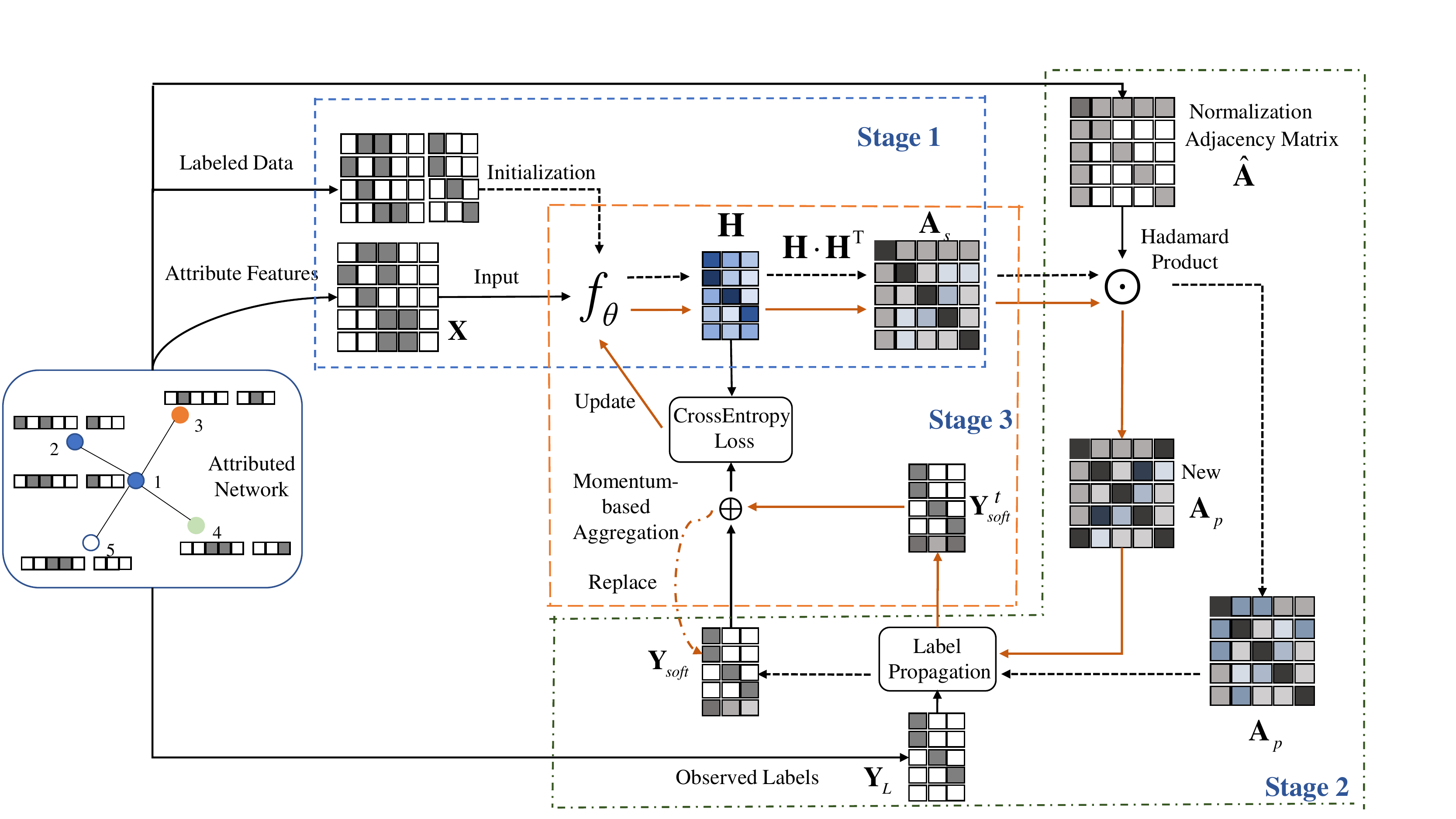}
	\caption{ 
	The framework of \name. It consists of three stages. In Stage 1, \name generates the similarity mask of nodes based on the attribute features and the labeled data. In Stage 2, \name integrates the similarity mask into the adjacency matrix by Hadamard product to create the propagation matrix, then generates the soft labels based on the observed labels and propagation adjacency matrix using label propagation. In Stage 3, \name develops an iterative refinement mechanism to alternately update the similarity mask and soft labels for improving the training performance.}
	\label{fig1}
\end{figure}

The goal of our Propagation with Adaptive Mask then Training (\name) is to train a superior neural network classifier $\boldsymbol{f}_{\theta}$ for node classification using the attribute features of nodes and the topology features of the graph. The framework of \name is shown in Figure \ref{fig1}. Specifically, \name contains three stages: calculate the attribute similarity mask, propagate labels, and train with an iterative refinement mechanism.

\subsection{Generating the Similarity Mask}
For the node classification task on attribute networks, the node attribute features are critical information. In general, nodes belonging to the same class tend to have similar attribute features. Thus, our \name utilizes the attribute information to generate an attribute similarity mask to capture the similarity between nodes. On the other hand, the attribute similarity mask can also refine the topological information based on attribute features to reduce the impact of structure noise during the propagation.

To guarantee a good quality of the similarity mask and keep simplicity on the model, we use a neural network classifier $\boldsymbol{f}_{\theta}$ (MLP in this paper) initialized with the labeled data to calculate the attribute similarity mask. For a randomly initialized $\boldsymbol{f}_{\theta}$, we first train on the labeled data with fixed training epochs. Then we get the representation matrix $\mathbf{H}$ of nodes through $\boldsymbol{f}_{\theta}$:
\begin{equation}
	\mathbf{H} = \boldsymbol{f}_{\theta}(\mathbf{X}). 
	\label{eq:h}
\end{equation}   
And the attribute similarity mask $\mathbf{A}_s$ is calculated as follows:
\begin{equation}
	\mathbf{A}_s = \mathbf{H}\mathbf{H}^{\top}.
	\label{eq:as}
\end{equation}

There are two motivations for this design: 1) after training with labeled data, $\boldsymbol{f}_{\theta}$ has a good classification performance, as shown in~\cite{DBLP:conf/www/DongCFHBD021}, so the output of $\boldsymbol{f}_{\theta}$ has the ability to ensure the quality of $\mathbf{A}_s$; 2) the goal of \name is to train a superior $\boldsymbol{f}_{\theta}$, thus initializing $\boldsymbol{f}_{\theta}$ with the labeled data to calculate $\mathbf{A}_s$ does not increase the model complexity as there is no additional parameter to be added to the model.

\subsection{Propagation with Similarity Mask}
The goal of stage 2 is to build a propagation matrix $\mathbf{A}_p$ that simultaneously preserves the attribute information and topology information of nodes, and propagate the observed labels on $\mathbf{A}_p$ using label propagation to create the soft labels for unlabeled nodes. 

For calculating $\mathbf{A}_p$, we combine $\mathbf{\hat{A}}$ and $\mathbf{A}_s$ using Hadamard product:
\begin{equation}
	\mathbf{A}_p = \hat{\mathbf{A}}\odot \mathbf{A}_s.
	\label{eq:ap}
\end{equation}

The meanings of the above operation are twofold. On the one hand, the elements of $\mathbf{\hat{A}}$ and $\mathbf{A}_s$ represent the topology similarity and attribute similarity of the node pairs, respectively. Thus the elements of $\mathbf{A}_p$ preserve these two types of similarity for node pairs, which enhances the representation ability of the adjacency matrix. On the other hand, $\mathbf{\hat{A}}$ only contains the first-order neighbors of nodes, so that using Hadamard product to combine $\mathbf{\hat{A}}$ and $\mathbf{A}_s$ could preserve the sparsity of $\mathbf{\hat{A}}$,  which will reduce the 
time complexity 
of the propagation operation. 

After the above process, we gain a weighted matrix $\mathbf{A}_p$. We further run a propagation operation with observed labels $\mathbf{Y}_{L}$ to generate the soft labels $\mathbf{Y}_{soft}$:
\begin{equation}
	\mathbf{Y}_{soft} = \bar{\mathbf{A}}_p \mathbf{Y}_{L}.
	\label{eq:ysoft}
\end{equation}

The propagation method we adopt is the approximate personalized PageRank for its high computational efficiency and superior performance~\cite{DBLP:conf/www/DongCFHBD021,DBLP:conf/iclr/KlicperaBG19}. $\bar{\mathbf{A}}_p$ represents the matrix of $\mathbf{A}_p$ after propagation, calculated as follows~\cite{DBLP:conf/www/DongCFHBD021}:
\begin{equation}
	\bar{\mathbf{A}}_p = (1-\alpha)^{K}\mathbf{A}_p^{K} + \alpha\sum_{k=0}^{K-1}(1-\alpha)^{k}\mathbf{A}_p^{k}, 
	\label{eq:propagation}
\end{equation}         
where $\alpha$ is the propagation bias and $K$ denotes the number of propagations. 

\subsection{Iterative Refinement Mechanism}
The goal of stage 3 is to train a superior $\boldsymbol{f}_{\theta}$ with \textbf{$\mathbf{Y}_{soft}$}. 
The objective function of \name is described as follows:
\begin{equation}
	\boldsymbol{L}_{\name}(\theta) = \boldsymbol{\ell}(\boldsymbol{f}_{\theta}(\mathbf{X}), \mathbf{Y}_{soft}),
	\label{eq:ptasm_loss}
\end{equation}           
where $\boldsymbol{\ell}(\cdot)$ indicates the cross entropy loss.

However, $\mathbf{Y}_{soft}$ is generated by using label propagation only once during the whole training phase, and  $\boldsymbol{f}_{\theta}$ can not be well trained because $\mathbf{Y}_{soft}$ could not precisely represent the soft labels of nodes. 
To tackle the above issue, inspired by some researches that do refinement iteratively in a mutual way~\cite{hicode,DBLP:conf/icml/XieGF16,DBLP:journals/corr/abs-2009-01674}, we develop an iterative refinement mechanism to improve the training 
performance of model $\boldsymbol{f}_{\theta}$. 
In this mechanism, $\boldsymbol{f}_{\theta}$ will be updated periodically when meeting the update condition that epoch $t$ meets $t\%t_{u}=0$ ( $t_{u}$ is a fixed number). 
Then, the new $\mathbf{A}_s$ can be calculated based on the updated $\boldsymbol{f}_{\theta}$ according to \autoref{eq:h} and \autoref{eq:as}. Finally, \name calculates $\mathbf{Y}^t_{soft}$ at epoch $t$ according to \autoref{eq:ap} and \autoref{eq:ysoft}. 
Differing from previous works~\cite{DBLP:conf/icml/XieGF16,DBLP:journals/corr/abs-2009-01674} that update $\mathbf{Y}{_{soft}}$ by setting $\mathbf{Y}{_{soft}} \leftarrow \mathbf{Y}^t_{soft}$, we perform a momentum-based strategy~\cite{he2020momentum} to update $\mathbf{Y}{_{soft}}$ to avoid instability and information loss:
\begin{equation}
	\mathbf{Y}{_{soft}} \leftarrow \beta \mathbf{Y}{_{soft}} + (1-\beta) \mathbf{Y}^t_{soft},
	\label{eq:yupdate}
\end{equation}  
where $\beta \in \lbrack 0,1)$ is the momentum coefficient to control the contribution to the label update. The overall learning algorithm of \name is summarized in Algorithm \ref{alg:Framwork}. 

The intuition is that the well-trained $\boldsymbol{f}_{\theta}$ will enhance the quality of $\mathbf{Y}{_{soft}}$, and high quality $\mathbf{Y}{_{soft}}$ will improve the training performance of $\boldsymbol{f}_{\theta}$. As the training process goes on, both $\boldsymbol{f}_{\theta}$ and $\mathbf{Y}{_{soft}}$ will be promoted.
Since the updates of $\boldsymbol{f}_{\theta}$ and $\mathbf{Y}{_{soft}}$ are iterative,
we call it an iterative refinement mechanism. 
In addition, the similarity mask $\mathbf{A}_s$ generated by $\boldsymbol{f}_{\theta}$ is changing during the training process. Hence, $\mathbf{A}_s$ could be regarded as an adaptive similarity mask.    

\begin{algorithm}[t]  
	\caption{The learning algorithm of \name }  
	\label{alg:Framwork}  
	\begin{algorithmic}[1]  
		\Require
		Adjacency matrix $\mathbf{A}$;
		Feature matrix $\mathbf{X}$;
		Observed labels $\mathbf{Y}_L$;
		\Ensure  
		Neural network classifier $\boldsymbol{f}_{\theta}$ ;
		\State Initialize $\boldsymbol{f}_{\theta}$ with labeled data;
		\State Generate the attribute similarity mask $\mathbf{A}_s$ by \autoref{eq:as};    
		\State Generate soft label ${\mathbf{Y}}_{soft}$ by \autoref{eq:ysoft};     
		\For{$t=1$ to $Epoch_{max}$}
		
		\If {$t \% t_u = 0$}
		\State Update $\mathbf{A}_s$ by \autoref{eq:as}
		\State Generate $\mathbf{Y}^t_{soft}$ by \autoref{eq:ysoft}
		\State Update ${\mathbf{Y}}_{soft}$ by \autoref{eq:yupdate}
		\EndIf
		\State Calculate loss $\boldsymbol{L}_{\name}(\theta)$ by \autoref{eq:ptasm_loss}
		\State Optimize $\theta$ by minimizing $\boldsymbol{L}_{\name}$ with gradient descent
		\EndFor 
		\label{code:fram:extract}  
		
		\label{code:fram:select} \\  
		\Return Neural network classifier $\boldsymbol{f}_{\theta}$;  
	\end{algorithmic}  
\end{algorithm}   

After the training process, we generate the final prediction using $\boldsymbol{f}_{\theta}$ based on \autoref{eq:PT_OUT} for predicting the labels of unlabeled nodes.

\subsection{Complexity Analysis}
The time complexity of \name could be analyzed through the two phases: the training phase and inference phase. 

Suppose $\boldsymbol{f}_{\theta}$ is an MLP with one hidden layer of dimension $f$. In the training phase, if the training epoch $t$ does not meet the update condition, the time complexity of each epoch will be $\Phi(\boldsymbol{f}_{\theta}(\mathbf{X}))=O(nf(d+c))$, where $\Phi(\boldsymbol{f}_{\theta}(\mathbf{X}))$ represents the complexity of operation $\boldsymbol{f}_{\theta}(\mathbf{X})$, $n$ represents the number of nodes, $c$ represents the number of labels and $d$ represents the dimension of raw attribute features; if $t$ meets the condition, the complexity of each epoch will be consist of four parts: $\Phi(\mathbf{H}\mathbf{H}^{\top})+ \Phi(\hat{\mathbf{A}} \odot \mathbf{A}_s) +\Phi(\mathbf{A}_p \mathbf{Y}_{V_L})+ \Phi(\boldsymbol{f}_{\theta}(\mathbf{X}))
=O(n^2(2c+1) + nf(d+c))$. 

In the inference phase, the complexity is $\Phi(\boldsymbol{f}_{\theta}(\mathbf{X})) +\Phi(\bar{\mathbf{A}}\mathbf{H}) = O(n^2c+nf(d+c))$. Sparse matrix techniques could be used for optimizing the above operations except $\boldsymbol{f}_{\theta}(\mathbf{X})$ and $\mathbf{H}\mathbf{H}^{\top}$. Since $\mathbf{H}$ is a dense matrix, the complexity of $\mathbf{H}\mathbf{H}^{\top}$ will be high when the number of nodes is large. And the similarity mask is created by $\mathbf{H}\mathbf{H}^{\top}$, thus an important future work is to optimize the method of generating the similarity mask.

\section{Experiments}
In this section, we first introduce the experimental setup, including benchmark datasets, baseline methods and implementation details. Then we conduct 
extensive experiments to answer the following questions:
 1) How does \name perform as compared with existing baselines in the semi-supervised node classification task?
 2) How does the robustness of \name to the graph structure noise, comparing with existing baselines?
 3) How do the key designs of \name help improve the model performance? 
 4) How do the key hyper-parameters affect the performance of \name?

\subsection{Experimental Setup}

\textbf{{\itshape Datasets}}. Following \cite{DBLP:conf/www/DongCFHBD021}, we choose four widely used attribute network datasets for experiments, including Cora\_ML~\cite{DBLP:journals/ir/McCallumNRS00}, Citeseer~\cite{DBLP:journals/aim/SenNBGGE08}, Pubmed~\cite{namata2012query} and Microsoft Academic~\cite{DBLP:journals/corr/abs-1811-05868}. Cora\_ML, Citeseer, Pubmed are citation networks and Microsoft Academic (MS\_ACA) is a co-authorship network. Nodes in these networks represent entities (\eg, papers in citation networks) and edges represent relationships between entities (\eg, citation relationships in citation networks). The statistics of datasets are reported in Table \ref{tab:dataset}. For the semi-supervised node classification setting, following \cite{DBLP:conf/www/DongCFHBD021}, we split each dataset into three partitions: training set, early-stopping set and test set. Each class has 20 labeled nodes in the training set. All the datasets are available at the website\footnote{https://github.com/DongHande} from \cite{DBLP:conf/www/DongCFHBD021}. 

\textbf{{\itshape Baselines}}. 
We choose the baselines from the following three categories: GCN-based methods, decoupled GCN-based methods and PT-based methods. For GCN-based methods, we select GCN~\cite{DBLP:conf/iclr/KipfW17}, GAT~\cite{DBLP:conf/iclr/VelickovicCCRLB18} and RGCN~\cite{DBLP:conf/kdd/ZhuZ0019}. For decoupled GCN-based methods, we select SGCN~\cite{DBLP:conf/icml/WuSZFYW19} and APPNP~\cite{DBLP:conf/iclr/KlicperaBG19}. For PT-based methods, we select PTS~\cite{DBLP:conf/www/DongCFHBD021} and PTA~\cite{DBLP:conf/www/DongCFHBD021}.  
For variants of \name, we use PAMT$_r$ to represent a variant of \name that uses a randomly initialized $\boldsymbol{f}_{\theta}$ to generate the similarity mask.

\textbf{{\itshape Implementation Details}}. The experiments were conducted on a server with 1 I9-9900k CPU, 1 RTX 2080TI GPU and 64G RAM.
The codes of all methods are based on Python 3.8 and Pytorch 1.8.0.
We use the Adam optimizer~\cite{DBLP:journals/corr/KingmaB14} for learning the parameters of all models. 
The hyper-parameters of the baselines use the recommendation settings in their official implementation. 
For \name, we determine the hyper-parameters by grid search.
The detailed hyper-parameter settings of \name on each dataset are summarized in Table~\ref{tab:hy-para}.
Here $dim$ denotes the dimension of hidden layer of $\boldsymbol{f}_{\theta}$; $wd$ and $lr$ denote the weight-decay coefficient and learning rate, respectively; $drop$ denotes the dropout rate.

\begin{table}[t]
    \centering
	\caption{Statistics on datasets.}
	\label{tab:dataset}
	\begin{tabular}{lrrrr}
		\toprule
		Dataset & \#Nodes & \#Edges & \#Features & \#Classes\\
		\midrule
		Cora\_ML & 2,810 & 7,981 & 2,879 & 7\\
		Citeseer & 2,110 & 3,668 & 3,703 & 6\\
		Pubmed & 19,717 & 44,324 & 500 & 3\\
		MS\_ACA & 18,333 & 81,894 & 6,805 & 15\\
		\bottomrule
	\end{tabular}
	
\end{table}

\begin{table}[t]
    \centering
	\caption{Hyper-parameters of \name on each dataset.}
	\label{tab:hy-para}
	\begin{tabular}{lcccccccc}
		\toprule
		Dataset & $dim$ & $\alpha$ & $wd$ & $lr$ & $\beta$ & $K$ & $drop$ & $t_{u}$\\
		\midrule
		Cora\_ML & 128 & 0.10 & 0.025 & 0.05 & 0.50 & 10 & 0.20 & 30 \\
		Citeseer & 128 & 0.15 & 0.055 & 0.10 & 0.25 & 10 & 0.15 & 20 \\
		Pubmed & 128 & 0.10 & 0.015 & 0.10 & 0.10 & 10 & 0.35 & 10 \\
		MS\_ACA & 256 & 0.10 & 0.010 & 0.05 & 0.10 & 10 & 0.35 & 10 \\
		\bottomrule
	\end{tabular}
\end{table}
 
\begin{table}[t]
    
    \centering
	\caption{Results of all models in terms of accuracy (\%) on different datasets. 
	The best results are in bold and the second bests are underlined.
	}
	\label{tab:rq1}

	\begin{tabular}{lcccc}
		\toprule
		Method  &Cora\_ML   &Citeseer   &Pubmed & MS\_ACA\\
		\midrule
		GCN &82.90 $\pm$ 0.41  & 73.56 $\pm$ 0.59 & 76.75 $\pm$ 0.63  & 91.26 $\pm$ 0.18\\
		GAT &83.46 $\pm$ 0.58  & 72.79 $\pm$ 0.24 & 78.97 $\pm$ 0.24  & 91.43 $\pm$ 0.05\\
		RGCN &84.18 $\pm$ 0.15  & 72.83 $\pm$ 0.15 & 78.77 $\pm$ 0.11  & 92.61 $\pm$ 0.08\\
		SGCN &77.44 $\pm$ 0.87  & \underline{75.89 $\pm$ 0.36} & 70.68 $\pm$ 1.50  & 90.83 $\pm$ 0.17\\
		APPNP &\underline{85.63 $\pm$ 0.18}  & 75.51 $\pm$ 0.35 & 78.86 $\pm$ 0.57  & 92.41 $\pm$ 0.19\\
		PTS &85.62 $\pm$ 0.37  & 75.74 $\pm$ 0.35 & 78.79 $\pm$ 0.69  & 91.85 $\pm$ 0.17\\
		PTA &85.50 $\pm$ 0.22  & 75.72 $\pm$ 0.41 & \underline{79.08 $\pm$ 0.77}  & \underline{92.72 $\pm$ 0.09}\\
		
		PAMT$_r$ &83.27 $\pm$ 0.84 & 69.01 $\pm$ 0.63 & 76.67 $\pm$ 0.56  & 92.35 $\pm$ 0.48\\
		\name &\textbf{86.01 $\pm$ 0.32} & \textbf{76.98 $\pm$ 0.24} & \textbf{79.38 $\pm$ 0.44}  & \textbf{92.92 $\pm$ 0.14}\\
		\bottomrule
	\end{tabular}
	
\end{table}

\subsection{Performance Comparison}
We first compare the performance of \name with the baselines for the semi-supervised node classification task. Table \ref{tab:rq1} reports the main results of all models in terms of accuracy and standard deviation with random data splits. 
In general, \name outperforms all the baselines on every datasets. Specifically, \name beats PT-based methods, which indicates that introducing the attribute features into the propagation during the training phase can improve the training performance. In addition, \name outperforms its variant PAMT$_r$, which indicates initializing $\boldsymbol{f}_{\theta}$ with the labeled data is helpful to increase the model performance.  

We also observe that, in general, the decoupled GCN-based methods (\eg APPNP and PTA) outperform GCN-based methods (\eg GCN and GAT) over all datasets. This is mainly because decoupled GCN-based methods can allow the information propagate deeper on networks than GCN-based methods and exhibits the superiority of the decoupled design.

\begin{figure}[t]
	\centering 
	\subfigure[Cora\_ML]{\includegraphics[width=2.2in]{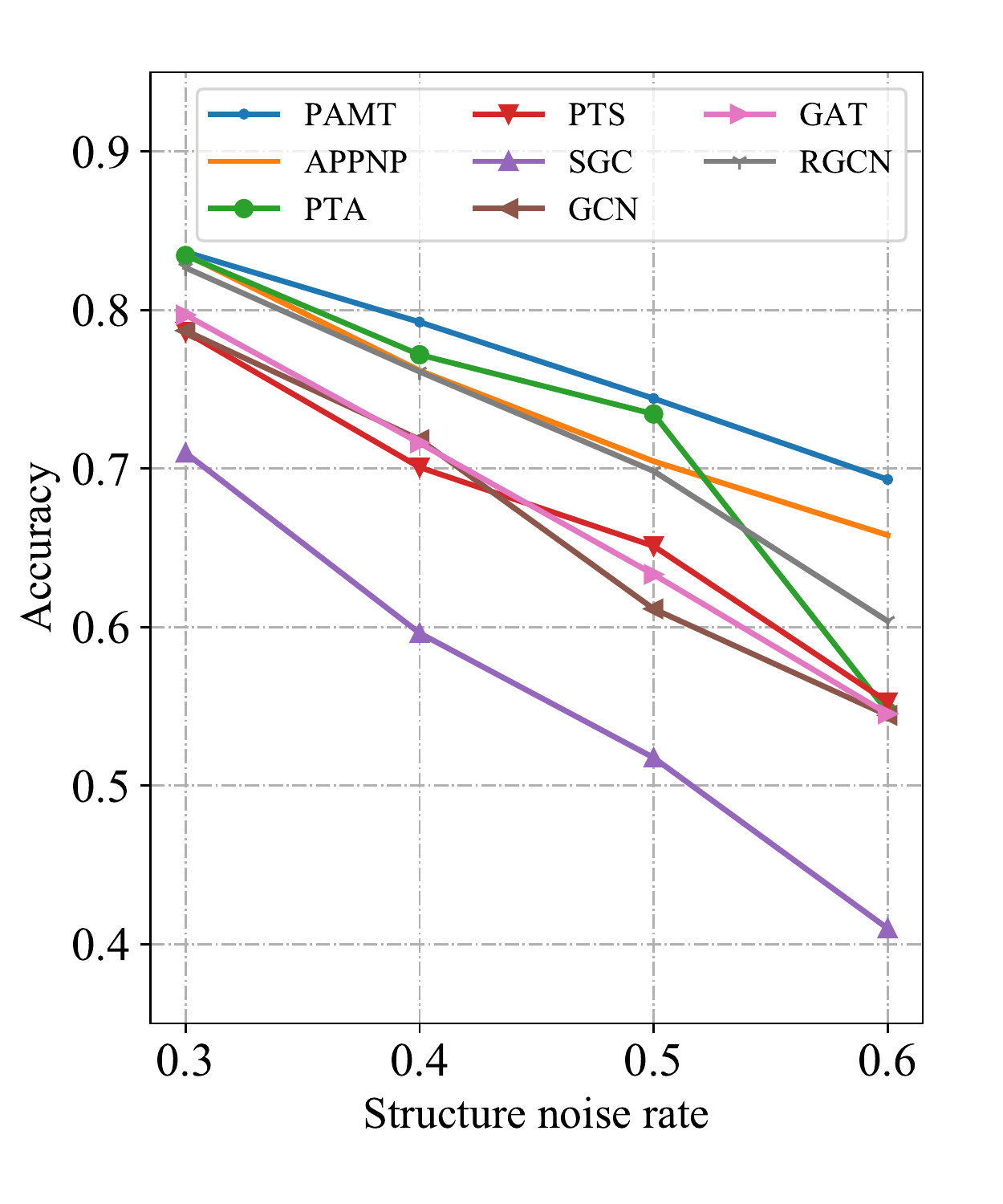}}
	\subfigure[Citeseer]{\includegraphics[width=2.2in]{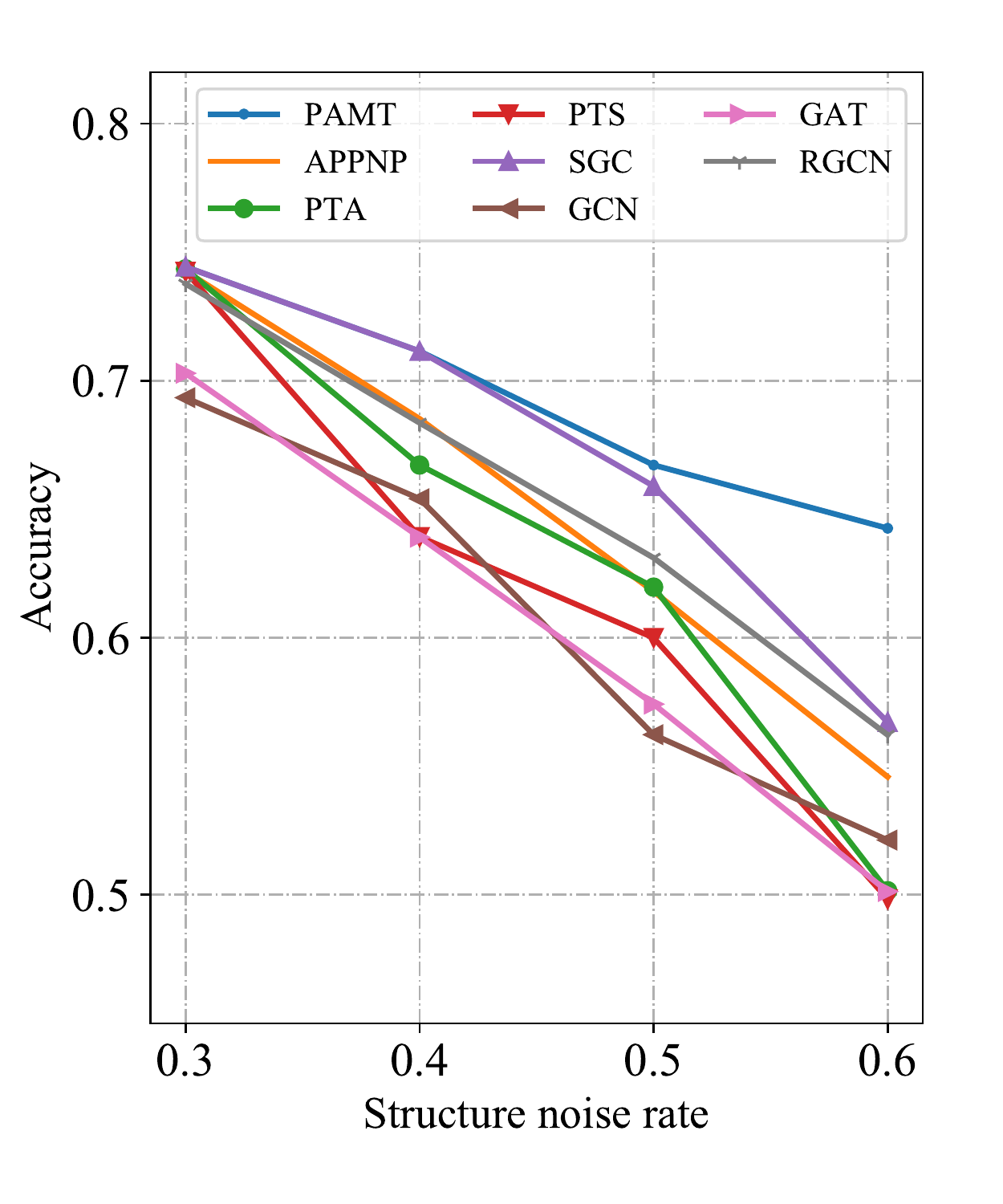}}
	\subfigure[Pubmed]{\includegraphics[width=2.2in]{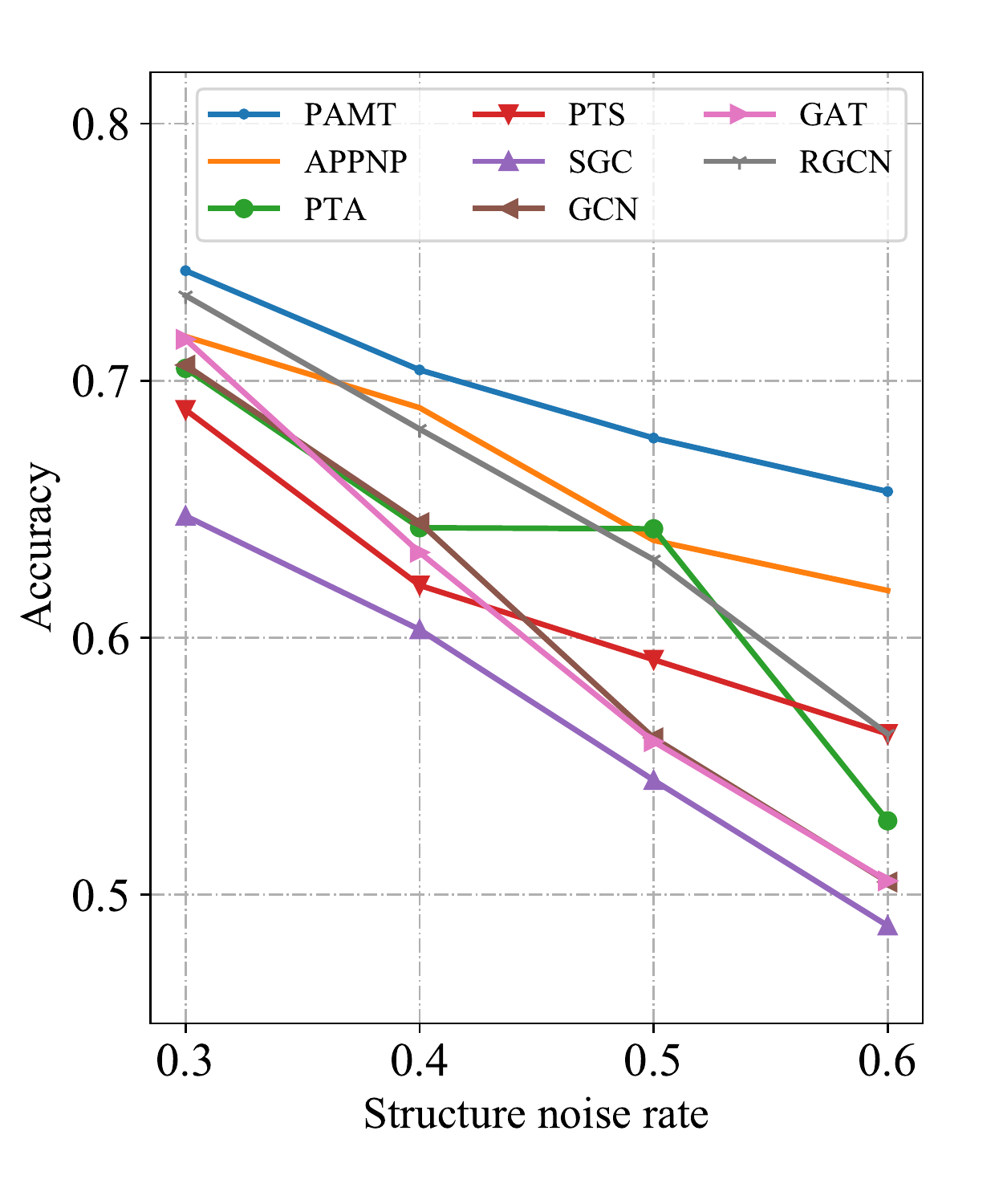}}
	\subfigure[MS\_ACA]{\includegraphics[width=2.2in]{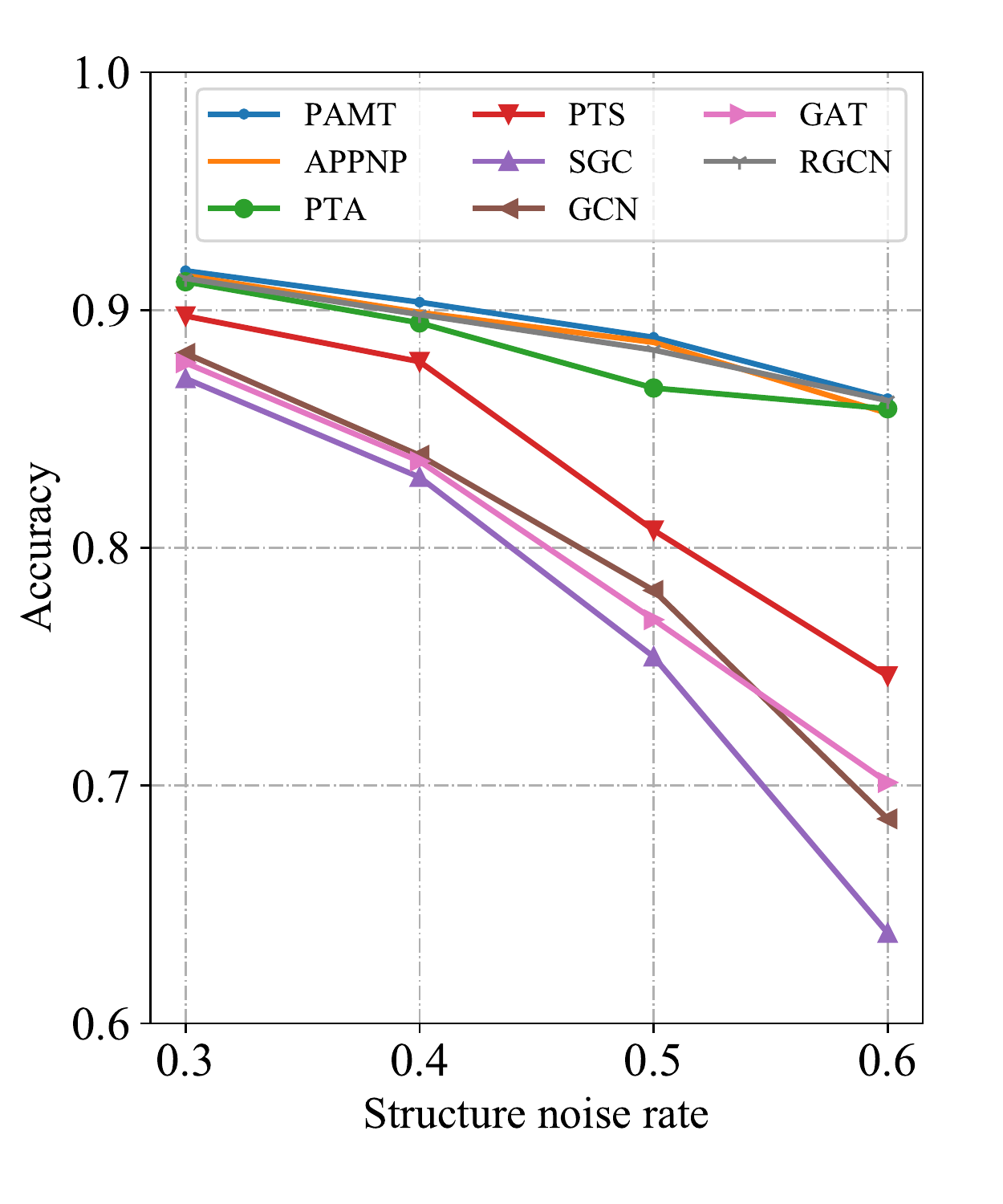}}
	\caption{Performance comparison 
	after adding different rates of structure noise.}
	\label{fig:exp2}
\end{figure}

\subsection{Comparison on Robustness to Structure Noise}
We conduct experiments to further study the model robustness to structure noise. 
As mentioned before, we regard the edge connecting nodes of different labels as structure noise.
So that for adding structure noise to the graph, we randomly remove the edge connecting nodes of the same label and add the edge connecting nodes of the different labels which is following the work of Dong \etal~\cite{DBLP:conf/www/DongCFHBD021}.
Our goal is to explore the performance of all models under different rates of structure noise. 
Since the original rates of structure noise over all datasets are less than 0.3, our experiments begin the structure noise rate at 0.3 and end at 0.6. 
Figure \ref{fig:exp2} reports the results, from which we have the following observations:

1) \name exhibits higher robustness to the structure noise, outperforming other methods on all the four datasets. 
These results validate that the design of introducing the attribute information into the propagation operation is beneficial to improve the model robustness. 
On MS\_ACA, the performance of \name, PTA, APPNP and RGCN are very close. This is mainly because MS\_ACA provides abundant attribute features (refer to Table \ref{tab:dataset}) which are enough to train a competitive classifier for prediction. 

2) Decoupled GCN-based methods exhibit higher robustness to structure noise than GCN-based methods. This may due to that decoupled GCN-based methods can utilize more information based on the propagation process.
In addition, APPNP and PTA outperform SGC on Cora\_ML, Pubmed and MS\_ACA. This is mainly because the propagation method of APPNP and PTA is the personalized PageRank which will retain their own features of the nodes. In this way, the influence of structure noise will be reduced.

3) GCN-based methods, like GCN and GAT, are sensitive to structure noise. This is because they obey the coupled design that limits the ability of deeply capturing the information of networks, and they only utilize information from the first-order neighbors, making them easily be disturbed by structure noise.

\begin{table}[t]
   
    \centering
	\caption{\name and its variants.}
	\label{tab:rq3_v}
	\begin{tabular}{lccc}
		\toprule
		Method & \tabincell{c}{Similarity mask-based  \\  propagation}&\tabincell{c}{Iterative   refinement\\  mechanism}&\tabincell{c}{Momentum-based \\ update strategy}\\
		\midrule
		PTS & $\times$ & $\times$ & $\times$  \\
		PAMT$_0$ & \checkmark  & $\times$  & $\times$   \\
		PAMT$_1$ & \checkmark & \checkmark &$\times$  \\	
		\name & \checkmark & \checkmark & \checkmark \\
		\bottomrule
	\end{tabular}
	
\end{table} 

\begin{table}[t]
    
    \centering
	\caption{Ablation study on \name. }
	\label{tab:rq3_r}

	\begin{tabular}{lcccc}
		\toprule
		Method  &Cora\_ML   &Citeseer   &Pubmed &MS\_ACA\\
		\midrule
		PTS &85.62 $\pm$ 0.37  & 75.74 $\pm$ 0.35 & 78.79 $\pm$ 0.69  & 91.85 $\pm$ 0.17\\
		PAMT$_0$ &85.52 $\pm$ 0.34  & 76.62 $\pm$ 0.32 & 78.97 $\pm$ 0.34  & 92.43 $\pm$ 0.38\\
		PAMT$_1$ &85.66 $\pm$ 0.36  & 76.82 $\pm$ 0.12 & 79.12 $\pm$ 0.29  & 92.60 $\pm$ 0.36\\	
		\name &86.01 $\pm$ 0.32 & 76.98 $\pm$ 0.24 & 79.38 $\pm$ 0.44  & 92.92 $\pm$ 0.14\\
		\bottomrule
	\end{tabular}
	
\end{table}

\subsection{Ablation Study}
In this subsection, we conduct experiments to validate the contributions of the key designs of \name to the model performance. As mentioned in Section \ref{sec3}, there are three key designs in \name: 
similarity mask-based propagation, iterative refinement mechanism and momentum-based update strategy. If we remove all these components, \name will degenerate to PTS. For validating the effectiveness of the above designs, we propose two variants of \name: PAMT$_0$ and PAMT$_1$. In PAMT$_0$, only the similarity mask-based propagation is preserved, while   PAMT$_1$ only removes the momentum-based update strategy of \name. Characteristics of \name and its variants are summarized in Table \ref{tab:rq3_v}. 

The results are reported in Table \ref{tab:rq3_r}, from which we have the following observations: 
1) In general, PAMT$_0$ and PAMT$_1$ outperform PTS, showing that both the similarity mask-based propagation and the  iterative refinement mechanism can help to train a high-quality neural network classifier.
2) PAMT$_1$ beats PAMT$_0$ on all datasets, exhibiting the importance of the iterative refinement mechanism. 
3) PAMT outperforms PAMT$_1$. The reason may be that the direct replacement strategy is used in PAMT$_1$, which may lead instability and cause information loss. This phenomenon also indicates that the momentum-based update strategy can improve the model performance.

\begin{figure}[t]
	\centering 
	\subfigure[On propagation step $K$]{\includegraphics[width=3in]{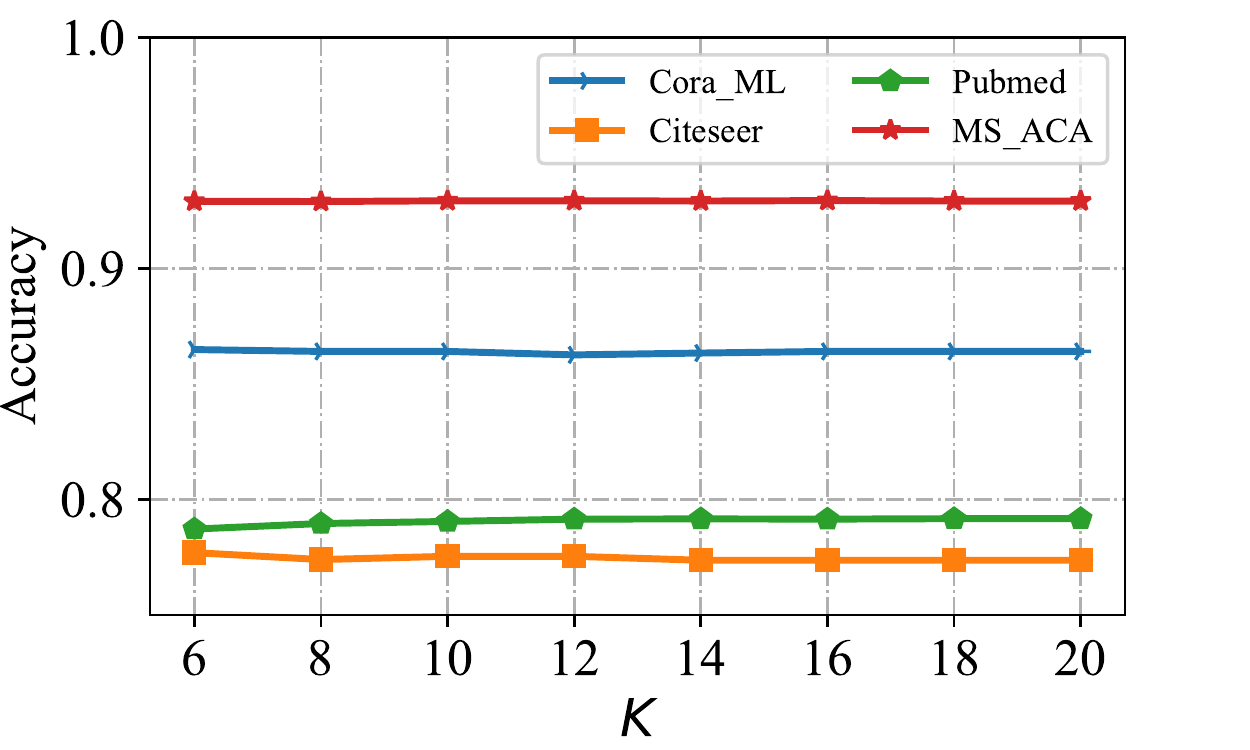}}
	\subfigure[On propagation bias $\alpha$]{\includegraphics[width=3in]{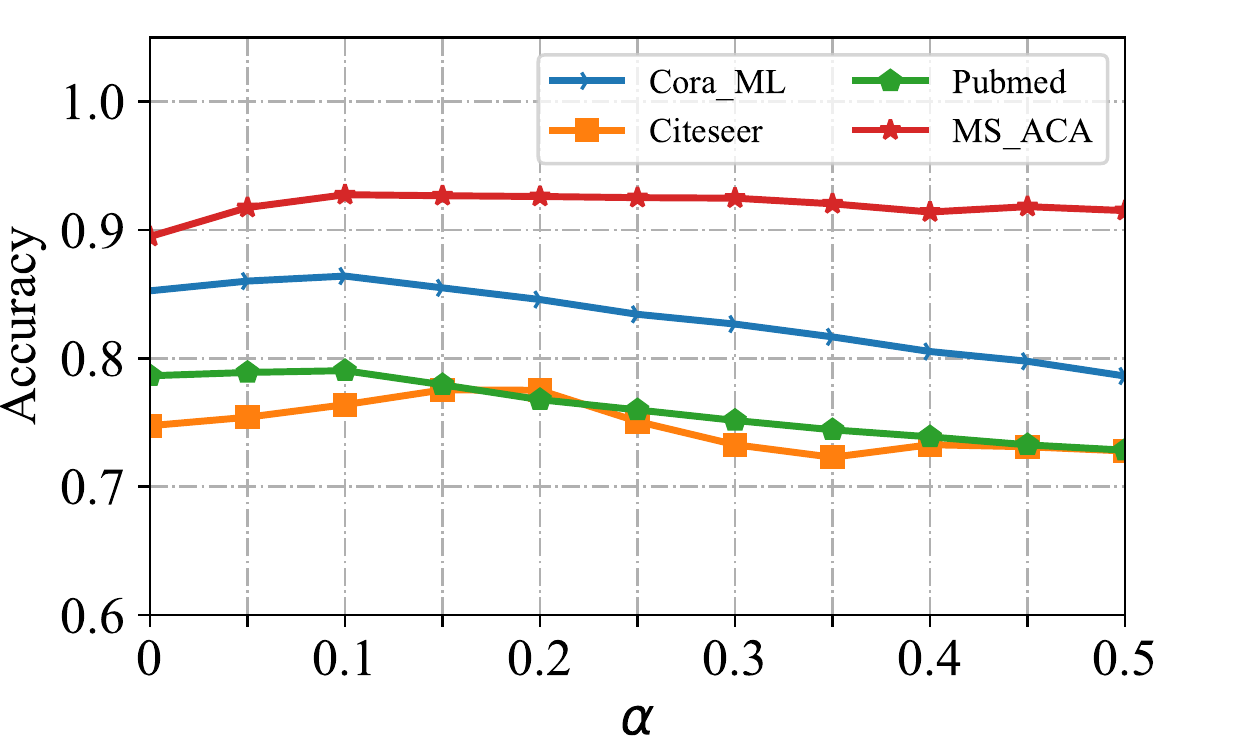}}
	\caption{ Performance under different parameters.}
	\label{fig:exp4}
\end{figure}

\subsection{Parameter Study}
We continue to conduct experiments to examine the sensitivity of the key parameters on the model performance, the propagation step $K$ and the propagation bias $\alpha$ in \autoref{eq:propagation}. $K$ represents the range of information that nodes can aggregate on the network and $\alpha$ means the proportion of information aggregated from neighbor nodes. These two parameters determine the quality of the propagation operation to a great extent and further affect the process of training $f_{\theta}$. 

To validate their influence on the model performance, we conduct the following experiments: 
1) Fix $\alpha$ and vary $K$ from $\{6, 8,...,20\}$ to observe the influence of $K$ to the model performance. Since the average lengths of the shortest paths of these datasets are greater than five~\cite{DBLP:conf/iclr/KlicperaBG19}, we set the first value of $K$ to 6. 
2) Fix $K$ and vary $\alpha$ from $\{0,0.05,0.1,..., 0.5\}$ to observe the impact of $\alpha$ to the model performance. 
The results are shown in Figure \ref{fig:exp4}.         

\textbf{{\itshape The influence of $K$}}. 
In general, the model performance varies slightly with the change of $K$, based on the results from Figure \ref{fig:exp4} (a). The results indicate that the information received by nodes will change less when $K$ reaches a certain value (\eg, 12 in Pubmed). It also shows that \name is not sensitive to the over-smoothing problem. 
In practice, we set $K=10$ for all datasets.
   
\textbf{{\itshape The influence of $\alpha$}}. From Figure \ref{fig:exp4} (b), we can see that the trend of the experimental results on each dataset is roughly the same. With the increment of $\alpha$, the model performance increases firstly and then declines. This phenomenon indicates that a small value of $\alpha$ helps the training process because the node will aggregate more information from its neighbors. In addition, $\alpha = 0$ mimics the traditional propagation of GCN-based methods (\eg, GCN and GAT), which demonstrates the benefits of propagating information with restarts. 
Based on the results of Figure \ref{fig:exp4} (b), we set $\alpha=0.1$ for Cora\_ML, Pubmed and MS\_ACA.
For Citeseer, we set $\alpha=0.15$.

\section{Conclusion}
In this paper, we propose a novel method called the {\itshape Propagation with Adaptive Mask then Training }(\name) for node classification on attributed networks in the semi-supervised 
setting. 
The key idea is that \name utilizes the attribute information to guide the propagation process in an iterative refinement mechanism on attributed networks. 
More specifically, \name uses the attribute features to generate the similarity mask of 
node pairs and further combines the similarity with the adjacency matrix to conduct the propagation matrix. During the training phase, \name updates the similarity mask by an iterative refinement mechanism. In this way, the influence of structure noise could be effectively reduced. Moreover, we develop a momentum-based update strategy to keep the training stable. The conducted experiments on four widely used attribute network datasets demonstrate that \name outperforms various advanced baselines for the semi-supervised node classification task and achieves higher robustness to structure noise.


\bibliographystyle{model1-num-names}

\bibliography{cas-refs}





\end{document}